%% file: main.tex
\def\year{2020}\relax
\documentclass[letterpaper]{article} 
\usepackage{aaai19}  
\usepackage{times}  
\usepackage{helvet} 
\usepackage{courier}  
\usepackage[hyphens]{url}  
\usepackage{graphicx} 
\usepackage{graphicx}  
\title{XferNAS: Transfer Neural Architecture Search}
\author{Martin Wistuba\textsuperscript{\rm }\\ 
\textsuperscript{\rm }IBM Research\\ 
martin.wistuba@ibm.com 
}

\usepackage[utf8]{inputenc} 
\usepackage[T1]{fontenc}    
\usepackage{url}            
\usepackage{booktabs}       
\usepackage{amsfonts}       
\usepackage{nicefrac}       
\usepackage{microtype}      
\usepackage{amsmath}
\usepackage{caption,subcaption} 
\usepackage{tikz}
\usetikzlibrary{positioning,calc,bayesnet}
\usepackage[numbers,sort&compress]{natbib}

\DeclareMathOperator*{\argmax}{arg\,max}
\DeclareMathOperator*{\argmin}{arg\,min}

\begin{document}

\maketitle

\begin{abstract}
  The term Neural Architecture Search (NAS) refers to the automatic optimization of network architectures for a new, previously unknown task.
  Since testing an architecture is computationally very expensive, many optimizers need days or even weeks to find suitable architectures.
  However, this search time can be significantly reduced if knowledge from previous searches on different tasks is reused.
  In this work, we propose a generally applicable framework that introduces only minor changes to existing optimizers to leverage this feature.
  As an example, we select an existing optimizer and demonstrate the complexity of the integration of the framework as well as its impact.
  In experiments on CIFAR-10 and CIFAR-100, we observe a reduction in the search time from 200 to only 6 GPU days, a speed up by a factor of 33.
  In addition, we observe new records of 1.99 and 14.06 for NAS optimizers on the CIFAR benchmarks, respectively.
  In a separate study, we analyze the impact of the amount of source and target data.
  Empirically, we demonstrate that the proposed framework generally gives better results and, in the worst case, is just as good as the unmodified optimizer.
\end{abstract}

\section{Introduction}
For most recent advances in machine learning ranging across a wide variety of applications (image recognition, natural language processing, autonomous driving), deep learning has been one of the key contributing technologies.
The search for an optimal deep learning architecture is of great practical relevance and is a tedious process that is often left to manual configuration.
Neural Architecture Search (NAS) is the umbrella term describing all methods that automate this search process.
Common optimization methods use techniques from reinforcement learning~\cite{Zoph2017_Neural,Baker2017_Designing,Cai2018_Efficient,Zoph2018_Learning,Zhong2018_Practical,Cai2018_Path} or evolutionary algorithms~\cite{Real2017_Large,Liu2018_Hierarchical,Real2019_Aging}, or are based on surrogate models~\cite{Liu2018_Progressive,Luo2018_NAO}.
The search is a computationally expensive task since it requires training hundreds or thousands of models, each of which requires few hours of training on a GPU~\cite{Zoph2017_Neural,Liu2018_Hierarchical,Zoph2018_Learning,Real2019_Aging,Liu2018_Progressive,Luo2018_NAO}.
All of these optimization methods have in common that they consider every new problem independently without considering previous experiences.
However, it is a common knowledge that well-performing architectures for one task can be transferred to other tasks and even achieve good performance.
Architectures discovered for CIFAR-10 have not only been transferred to CIFAR-100 and ImageNet~\cite{Zoph2018_Learning,Liu2018_Progressive,Real2019_Aging,Luo2018_NAO} but have also been transferred from object recognition to object detection tasks~\cite{Redmon2016_YOLO,Liu2016_SSD}.
This suggests that the response function for different tasks, i.e. an architecture-score mapping, shares commonalities.

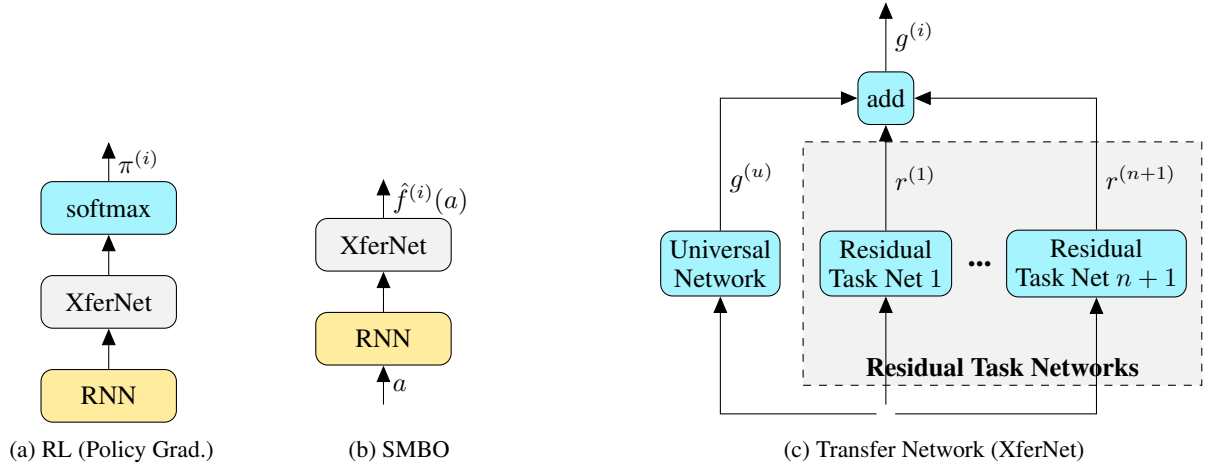
\begin{figure*}
  \centering
  \begin{subfigure}[b]{0.2\textwidth}
      \centering
      \input{xfernet-rl.tikz}
      \caption{RL (Policy Grad.)}
      \label{fig:transfer-rl}
  \end{subfigure}
  \hfill
  \begin{subfigure}[b]{0.2\textwidth}
      \centering
      \input{xfernet-smbo.tikz}
      \caption{SMBO}
      \label{fig:transfer-smbo}
  \end{subfigure}
  \hfill
  \begin{subfigure}[b]{0.57\textwidth}
      \centering
      \input{xfernet.tikz}
      \caption{Transfer Network (XferNet)}
      \label{fig:xfernet}
  \end{subfigure}
  \caption{An example of the integration of the transfer network into RL-based (a) and surrogate model-based~(b) optimizers.
   The transfer network (c) unravels independent and task-specific influences.}
  \label{fig:general-framework}
\end{figure*}

The central idea in this work lies in the development of a search method that uses knowledge acquired across previously explored tasks to speed up the search for a new task.
For this we assume that the response functions can be decomposed into two parts: a universal part, which is shared across all tasks, and a task-specific part (Figure~\ref{fig:general-framework}).
We model these two functions with neural networks and determine their parameters with the help of the collected knowledge of architectures on different tasks.
This allows the search for a new task to start with the universal representation and only later learn and benefit from the task-specific representation.
This reduces the search time without negatively affecting the final solution.

The contributions in this paper are threefold:
\begin{itemize}
 \item First, we propose a general, minimally invasive framework that allows existing NAS optimizers to leverage knowledge from other data sets, e.g. obtained during previous searches.
 \item Second, as an example, we apply the framework to NAO~\cite{Luo2018_NAO}, a recent NAS optimizer, and derive XferNAS.
 This exemplifies the simplicity and elegance of extending existing NAS methods within our framework.
 \item Finally, we demonstrate the utility of the adapted optimizer XferNAS by searching for an architecture on CIFAR-10~\cite{Krizhevsky2009_CIFAR}.
 In only 6 GPU days (NAO needed 200 GPU days) we discover a new architecture with improved performance compared to the state-of-the-art methods.
 We confirm the transferability of the discovered architecture by applying it, unchanged, to CIFAR-100.
\end{itemize}

\section{Transfer Neural Architecture Search}
In this section, we introduce our general, minimally invasive framework for NAS optimizers to leverage knowledge from other data sets, e.g. obtained during previous searches.
First, we formally define the NAS problem and introduce our notation.
Then we motivate our approach and introduce the framework.
Finally, using the example of NAO~\cite{Luo2018_NAO}, we show what steps are required to integrate the framework with existing optimizers.
In this step, we derive XferNAS, which we examine in more detail in Section~\ref{sec:experiments}.

\subsection{Problem definition}

We define a general deep learning algorithm $\mathbb{L}$ as a mapping from the space of data sets $D$ and architectures $A$ to the space of models $M$,
\begin{equation}
    \mathbb{L}\ :\ D\times A\rightarrow M\ .
\end{equation}
For any given data set $d\in D$ and architecture $a\in A$, this mapping returns the solution to the standard machine learning problem that consists in minimizing a regularized loss function $\mathcal{L}$ with respect to the model parameters $\theta$ of architecture $a$ using the data $d$,
\begin{equation}
    \mathbb{L}\left(a,d\right) = \argmin_{m^{(a,\theta)}\in M^{(a)}}\,\mathcal{L}\left(m^{(a,\theta)}, d^{(\text{train})}\right) + \mathcal{R}\left(\theta\right)\ .
\end{equation}
Neural Architecture Search solves the following nested optimization problem:
given a data set $d$ and the search space $A$, find the optimal architecture $a^{\star}\in A$ which maximizes the objective function~$\mathcal{O}$ (defined by classification accuracy in the scope of this work) on the validation data,
\begin{equation}
     a^\star = \argmax_{a\in A}\, \mathcal{O}\left(\mathbb{L}\left(a,d^{(\text{train})}\right),d^{(\text{valid})}\right)=\argmax_{a\in A} f\left(a\right)\ .\label{eq:NAS-problem}
\end{equation}
Thus, Neural Architecture Search can be considered a global black-box optimization problem where the aim is to maximize the \emph{response function} $f$.
It is worth noting that the evaluation of $f$ at any point $a$ is computationally expensive since it involves training and evaluating a deep learning model.

In this work we assume that we have access to knowledge about the response functions on some \emph{source tasks} $1,\ldots,n$, referred to as \emph{source knowledge}.
The idea is to leverage the source knowledge to address the NAS problem defined in Equation~\eqref{eq:NAS-problem} for a new task, the \emph{target task} $n+1$.
By no means, the source knowledge is sufficient to yield an optimal architecture for the new task.
Sample architectures must be evaluated on the target task in order to gain knowledge about the target response function.
We call the knowledge accumulated in this process the \emph{target knowledge} and refer to the combined source and target knowledge as \emph{observation history}.
In the context of this work, we refrain from transferring model weights.

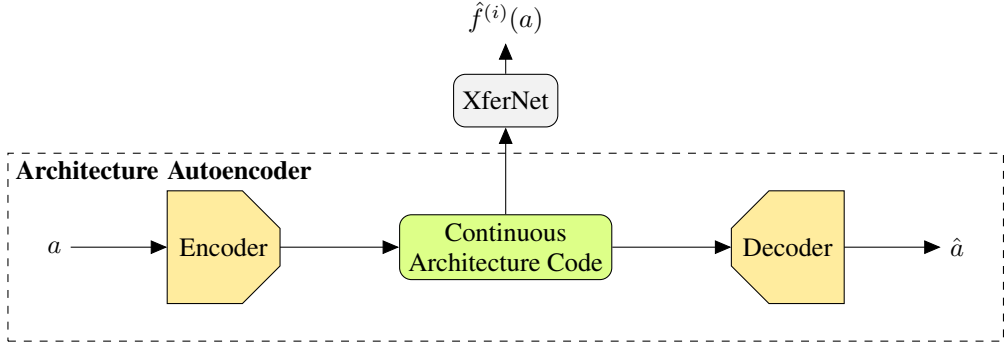
\begin{figure*}
  \centering
  \input{optimizer-framework.tikz}
  \caption{XferNAS: Integration of the transfer network in NAO.}\label{fig:optimizer-framework}
\end{figure*}
\subsection{Transfer network}\label{sub:xfernet}
The most common NAS optimizers are based on reinforcement learning (RL)~\cite{Zoph2017_Neural,Baker2017_Designing,Cai2018_Efficient,Zoph2018_Learning,Zhong2018_Practical,Cai2018_Path} or surrogate model-based optimization (SMBO)~\cite{Liu2018_Progressive,Luo2018_NAO}.
Many RL-based methods are based on policy gradient methods and use a controller (a neural network) to provide a distribution over architectures.
The controller is optimized in order to learn a policy $\pi$ which maximizes the response function of the target task.
Alternatively, SMBO methods use a surrogate model $\hat{f}^{(n+1)}$ to approximate the target response function $f^{(n+1)}$.
Both of these approaches rely on the feedback gathered by evaluating the target response function for several architectures.

We propose a general framework to transfer the knowledge gathered in previous experiments in order to speed up the search on the target task.
Current state-of-the-art NAS optimizers directly learn a task-dependent policy (RL) or a task-dependent surrogate model (SMBO) $g^{(i)}$.
The core idea in this work is to disentangle the contribution of the \emph{universal function} $g^{(u)}$ from the \emph{task-dependent function} $g^{(i)}$ by assuming
\begin{equation}
 g^{(i)}=g^{(u)}+r^{(i)}\ ,
\end{equation}
where $r^{(i)}$ is a \emph{task-dependent residual}.
This disentanglement is achieved by learning all parameters jointly on the observation history, where the universal function is included for all tasks while the task-dependent residual included only for its corresponding task.
The universal function can be interpreted as a function which models a good average solution across problems, whereas the task-dependent function is optimal for a particular task $i$.
The advantage is that we can warmstart any NAS optimizer for the target task where $r^{(n+1)}$ is unknown by using only $g^{(u)}$.
As soon as target knowledge is obtained, we can learn $r^{(n+1)}$ which enables us to benefit from the warmstart in the initial phase of the search and subsequently from the original NAS optimizer.
We sketch the idea of this transfer network in Figure~\ref{fig:general-framework} and provide an example for both, reinforcement learning and surrogate model-based optimizers.
The functions $g^{(u)}$ and $r^{(i)}$ are modeled by a neural network, referred to as the universal network and the residual task networks, respectively.
In the following, we exemplify this integration with the case of NAO~\cite{Luo2018_NAO} to provide a deeper understanding.

\subsection{XferNAS}\label{sub:xfernas}
In principle any of the existing network-based optimizers (RL or SMBO) can be easily extended with our proposed transfer network in order to leverage the source knowledge.
We demonstrate the use of the transfer network using the example of NAO~\cite{Luo2018_NAO}, one of the state-of-the-art optimizers.
NAO is based on two components, an auto-encoder and a performance predictor.
The auto-encoder first transforms the architecture encoding into a continuous architecture code by an encoder and then reconstructs the original encoding using a decoder.
The performance predictor predicts the accuracy on the validation split for a given architecture code.
XferNAS extends this architecture by integrating the transfer network into the performance predictor, which not only predicts accuracy for the target task, but also for all source tasks (Figure~\ref{fig:optimizer-framework}).
Here, the different prediction functions for each task $i$ are divided into a universal prediction function and a task-specific residual,
\begin{equation}
 \hat{f}^{(i)}=\hat{f}^{(u)}+r^{(i)}\ .
\end{equation}
In this case, the universal prediction function can be interpreted as a prediction function that models the general architecture bias, regardless of the task.
This is suitable for cases where there is no knowledge about the target task, as is the case at the beginning of a new search.
The task-specific residual models the task-specific peculiarities.
If knowledge about the task exists, it can be used to correct the prediction function for certain architecture codes.

The loss function to be optimized is identical to the original version of NAO and is a combination of the prediction loss function $L_{pred}$ and the reconstruction loss function $L_{rec}$,
\begin{equation}
\label{eq:loss}
L=\alpha L_{pred} + (1-\alpha) L_{rec}\ .
\end{equation}
However, the prediction loss function considers all observations to train the model for all tasks,
\begin{equation}
L_{pred} = \sum_{i=1}^{n+1}\sum_{a\in H^{(i)}} \left(f^{(i)}\left(a\right)-\hat{f}^{(i)}\left(a\right)\right)^2\ .
\end{equation}
$H^{(i)}$ is the set of all architectures which are evaluated for a task $i$ and for which the value of the response function is known.
The joint optimization of both loss functions guarantees that architectures which are close in the architecture code space exhibit similar behavior across tasks.

Once the model has been trained by minimizing the loss function, potentially better architectures can be determined for the target task.
The architecture codes of models with satisfactory performance serve as starting points for the optimization process.
Gradient-based optimization is used to modify the current architecture code to maximize the prediction function of the target task
\begin{equation}
\label{eq:gradient-search}
z\leftarrow z + \eta \frac{\partial\hat{f}^{(n+1)}}{\partial z}\ ,
\end{equation}
where $z$ is the current architecture code and $\eta$ is the step size.
The architecture encoding is reconstructed by applying the decoder on the final architecture code.
The step size is chosen large enough to get a new architecture, which is then evaluated on the target task.

XferNAS has two different phases.
In the first phase, the system lacks target knowledge and relies solely on the source knowledge.
The architectures with the highest accuracy on the source tasks serve as starting points for the determination of new candidates.
This is achieved by means of the process described in Equation~\eqref{eq:gradient-search} with $\eta=10$.
Having accumulated some target knowledge, the second phase selects as starting points the models with high accuracy on the target task.
To keep the search time low, we only examine 33 architectures.
Details of data splits, model training and used hardware are provided in the appendix.

\paragraph{Implementation details}
The transfer network has been integrated into the publicly available code of NAO\footnote{\url{https://github.com/renqianluo/NAO}}, allowing a fair comparison to Luo et al.~\cite{Luo2018_NAO}.
In our experiments we retain the prescribed architecture and its hyperparameters.
However, for the sake of completeness, we repeat this description.
LSTM models are used to model the encoder and decoder.
The encoder uses an embedding size of 32 and a hidden state size of 96, whereas the decoder uses a hidden state size of 96 in combination with an attention mechanism.
Mean pooling is applied to the encoder LSTM's output to obtain a vector which serves as the input to the transfer network.
The universal and the residual task networks are modelled with feed forward neural networks.
Adam~\cite{Kingma2015_Adam} is used to minimize Equation~\eqref{eq:loss} with learning rate set to $10^{-3}$, trade-off parameter $\alpha$ to 0.8, and weight decay to $10^{-4}$.

\section{Related work}
Neural Architecture Search (NAS), the structural optimization of neural networks, is solved with a variety of optimization techniques.
These include reinforcement learning~\cite{Zoph2017_Neural,Baker2017_Designing,Cai2018_Efficient,Zoph2018_Learning,Zhong2018_Practical,Cai2018_Path,Wistuba2018_Practical}, evolutionary algorithms~\cite{Real2017_Large,Liu2018_Hierarchical,Real2019_Aging,Wistuba2018_Deep}, and surrogate model-based optimization~\cite{Liu2018_Progressive,Luo2018_NAO}.
These techniques have made great advancements with the idea of sharing weights across different architectures which are sampled during the search process~\cite{Pham2018_ENAS,Bender2018_Understanding,Liu2018_DARTS,Xie2019_SNAS,Cai2019_Proxyless} instead of training them from scratch.
For a detailed overview we refer to a recent survey~\cite{Wistuba2019_A}.

A new but promising idea is to transfer knowledge from previous search processes to new ones \cite{Istrate2019_TAPAS,Wong2018_Transfer,Wistuba2019_Inductive}, which is analogous to the behavior of human experts.
We briefly discuss the current work in NAS for convolutional neural networks (CNNs) in this context.
TAPAS~\cite{Istrate2019_TAPAS} is an algorithm that starts with a simple architecture and extends it based on a prediction model.
For the predictions, first a very simple network is trained on the target data set.
Subsequently, the validation error is used to determine the similarity to previously examined data sets.
Based on this similarity, predictions of the validation error of different architectures on the target data set are obtained.
By means of these, a set of promising architectures are determined and evaluated on the target data set.
However, the prediction model is not able to leverage the additional information collected on the target data set.
T-NAML~\cite{Wong2018_Transfer} seeks to achieve the same effect without searching for new architectures.
Instead, it chooses a network which has been pre-trained on ImageNet and makes various decisions to adapt it to the target data set.
For this purpose, it uses a reinforcement learning method, which learns to optimize neural architectures across several data sets simultaneously.

\section{Experiments}\label{sec:experiments}

In this section we empirically evaluate XferNAS and compare the discovered architectures to the state-of-the-art.
Furthermore, we investigate the transferability of the discovered architectures by training it on a different data set without introducing any further changes.
In our final ablation study, we investigate the impact of amount of source and target data on the surrogate model's predictions.

\subsection{Architecture search space}\label{sub:search-space}
In our experiments, we use the widely adopted NASNet search space~\cite{Zoph2018_Learning}, which is also used by most optimizers that we compare with.
Architectures in this search space are based on two types of cells, normal cells and reduction cells.
These cells are combined to form the network architecture, with repeating units comprising of N normal cells followed by a reduction cell.
The sequence is repeated several times, doubling the number of filters each time.
The difference of the reduction cell from the normal cell is that it halves the dimension of feature maps.
Each cell consists of B blocks and the first sequence of N normal cells uses F filters.
The output of each block is the sum of the result of two operations, where the choice of operation and its input is the task of the optimizer.
There are \#op different operations, the input can be the output of each of the previous blocks in the same cell or the output of the previous two cells.
The output of a cell is defined by concatenating all block outputs that do not serve as input to at least one block.
The considered 19 operations are: identity, convolution ($1\times1$, $3\times3$, $1\times3+3\times1$, $1\times7+7\times1$), max/average pooling ($2\times2$, $3\times3$, $5\times5$), max pooling ($7\times7$), min pooling ($2\times2$), (dilated) separable convolution ($3\times3$, $5\times5$, $7\times7$).

\subsection{Source tasks}\label{sub:source-tasks}
Image recognition on Fashion-MNIST~\cite{Xiao2017_Fashion-MNIST}, Quickdraw~\cite{Ha2018_Quickdraw}, CIFAR-100~\cite{Krizhevsky2009_CIFAR} and SVHN~\cite{Netzer2011_SVHN} forms our four source tasks.
For each of these data sets, we evaluated 200 random architectures, giving us a total of 800 different architectures.
Every architecture is trained for 100 epochs with the settings described in the appendix.
Each of these architectures was trained on exactly one data set and none of these architectures were evaluated on the target task before or during the search.
Therefore, it is valid to conclude that the architecture found is new.

\begin{table*}
\centering
\small
\begin{tabular}{lcccccccc}
\toprule
Model & B & N & F & \#op & Error (\%) & \#params & M & GPU Days\\
\midrule
DenseNet-BC~\cite{Huang2017_Densely} & / & 100 & 40 & / & 3.46 & 25.6M & / & / \\
ResNeXt-29~\cite{Xie2017_Aggregated} & / & / & / & / & 3.58 & 68.1M & / & / \\
\midrule
NASNet-A~\cite{Zoph2018_Learning} & 5 & 6 & 32 & 13 & 3.41 & 3.3M & 20000 & 2000 \\
NASNet-B~\cite{Zoph2018_Learning} & 5 & 4 & N/A & 13& 3.73 & 2.6M & 20000 & 2000 \\
NASNet-C~\cite{Zoph2018_Learning} & 5 & 4 & N/A & 13 & 3.59 & 3.1M & 20000 & 2000 \\
Hier-EA~\cite{Liu2018_Hierarchical} & 5 & 2 & 64 & 6& 3.75 & 15.7M & 7000 & 300 \\
AmoebaNet-A~\cite{Real2019_Aging} & 5 & 6 & 36 & 10 & 3.34 & 3.2M & 20000 & 3150 \\
AmoebaNet-B~\cite{Real2019_Aging} & 5 & 6 & 36 & 19 & 3.37 & 2.8M & 27000 & 3150 \\
AmoebaNet-B~\cite{Real2019_Aging} & 5 & 6 & 80 & 19 & 3.04 & 13.7M & 27000 & 3150 \\
AmoebaNet-B~\cite{Real2019_Aging} & 5 & 6 & 128 & 19 & 2.98 & 34.9M & 27000 & 3150\\
AmoebaNet-B + Cutout~\cite{Real2019_Aging} & 5 & 6 & 128 & 19 & 2.13 & 34.9M & 27000 & 3150 \\
PNAS~\cite{Liu2018_Progressive} & 5 & 3 & 48 & 8 & 3.41 & 3.2M & 1280 & 225 \\ 
ENAS~\cite{Pham2018_ENAS} & 5 & 5 & 36 & 5 & 3.54 & 4.6M & / & 0.45 \\
Random-WS~\cite{Pham2018_ENAS} &5& 5 & 36 & 5 & 3.92 & 3.9M & / & 0.25\\
DARTS + Cutout~\cite{Liu2018_DARTS} & 5 & 6 & 36 & 7 & 2.83 & 4.6M & / & 4 \\
SNAS + Cutout~\cite{Xie2019_SNAS} & 5 & 6 & 36 & 7 & 2.85 & 2.8M & / & 1.5 \\
NAONet~\cite{Luo2018_NAO} & 5 & 6 & 36 & 11 & 3.18 & 10.6M & 1000 & 200 \\
NAONet~\cite{Luo2018_NAO} & 5 & 6 & 64 & 11 & 2.98 & 28.6M & 1000 & 200\\ 
NAONet + Cutout~\cite{Luo2018_NAO} & 5 & 6 & 128 & 11 & 2.11 & 128M & 1000 & 200\\ 
NAONet-WS~\cite{Luo2018_NAO} & 5 & 5 & 36 & 5 & 3.53 & 2.5M & / &0.3\\
\midrule
TAPAS~\cite{Istrate2019_TAPAS} & / & / & / & / & 6.33 & 2.7M & 1 & 0 \\
T-NAML~\cite{Wong2018_Transfer} & / & / & / & / & 3.5 & N/A & 150 & N/A \\
Best on source task CIFAR-100 & 5 & 6 & 32 & 19 & 4.14 & 6.1M & 200 & / \\
\midrule
XferNASNet & 5 & 6 & 32 & 19 & 3.37 & 4.5M & 33 & 6 \\
XferNASNet + Cutout & 5 & 6 & 32 & 19 & 2.70 & 4.5M & 33 & 6 \\
XferNASNet & 5 & 6 & 64 & 19 & 3.11 & 17.5M & 33 & 6 \\
XferNASNet + Cutout & 5 & 6 & 64 & 19 & 2.19 & 17.5M & 33 & 6 \\
XferNASNet & 5 & 6 & 128 & 19 & 2.88 & 69.5M & 33 & 6 \\
XferNASNet + Cutout & 5 & 6 & 128 & 19 & 1.99 & 69.5M & 33 & 6 \\
\bottomrule
\end{tabular}
\caption{
Classification error of discovered CNN models on CIFAR-10.
We denote the total number of models trained during the search by M.
B is the number of blocks, N the number of cells and F the number of filters.
\#op is the number of different operations considered in the cell which is an indicator of the search space complexity.
For more details on the hyperparameters \#op, B, N and F we refer to Section~\ref{sub:search-space}.
We expand the results collected by \cite{Luo2018_NAO} by the most recent works.
}
\label{tab:experiment-cifar10}
\end{table*}
\subsection{Image recognition on CIFAR-10}
We evaluate the proposed transfer framework using the CIFAR-10 benchmark data set and present the results in Table~\ref{tab:experiment-cifar10}.
The table is divided into four parts.
In the first part we list the results that some manually created architectures achieve.
In the second and third part we tabulate the results achieved by traditional NAS methods as well as those which transfer knowledge across tasks.
In the last part we list our results.
In contrast to some of the other search methods, we refrained from additional hyperparameter optimization of our final architecture (XferNASNet).

XferNAS is the extended version of NAO which additionally uses the transfer network; so this comparison is of particular interest.
We not only observe a significant drop in the search effort (number of evaluated models reduced from 1,000 to 33, search time reduced from 200 GPU days to 6), but also on the error obtained on the test set.
The smallest version of NAONet performs slightly better than XferNASNet (3.18 vs. 3.37), but also uses twice as many parameters.
If the data augmentation technique cutout~\cite{DeVries2018_Improved} is used, this minimal improvement turns around (2.11 vs. 1.99).

The transfer method TAPAS achieves significantly poor results (6.33 vs. 3.92 of the next better method).
The other transfer method T-NAML achieves an error rate of 3.5 which is not better than XferNASNet (3.37).
It should also be noted that T-NAML finetunes architectures which have been pre-trained on ImageNet.
Thus, not only the number of parameters is probably much higher, more data is used and no new architectures are found.
Therefore, it arguably solves a different task.
A very simple baseline is to select the best architecture on the most similar source task (CIFAR-100).
Objectively this baseline performs quite well (4.14) but compares poorly to XferNASNet.

XferNASNet performs very well compared to current gradient-based optimization methods such as DARTS and SNAS (2.70 versus 2.83 and 2.85, respectively).
It also provides good results compared to architectures such as NASNet or AmoebaNet which were discovered by time-consuming optimization methods (3.37 in 2,000 GPU days or 3.34 in 3,150 GPU days versus 3.41 in 6 GPU days).
\begin{table*}
\small
\centering
\begin{tabular}{lcccccccc}
\toprule
Model & B & N & F & \#op & Error (\%) & \#params \\
\midrule
DenseNet-BC~\cite{Huang2017_Densely} &/ &100 &40 &/ &17.18 &25.6M \\
Shake-shake~\cite{Gastaldi2017_Shake} &/ &/ &/ &/ &15.85 &34.4M \\
Shake-shake + Cutout~\cite{DeVries2018_Improved} &/ &/ &/ &/ &15.20 &34.4M \\
\midrule
NASNet-A~\cite{Zoph2018_Learning} & 5 & 6 & 32 & 13 & 19.70 & 3.3M \\
NASNet-A + Cutout~\cite{Zoph2018_Learning} & 5 & 6 & 32 & 13 & 16.58 & 3.3M \\
NASNet-A + Cutout~\cite{Zoph2018_Learning} & 5 & 6 & 128 & 13 & 16.03 & 50.9M \\
PNAS~\cite{Liu2018_Progressive} & 5 & 3 & 48 & 8 & 19.53 & 3.2M \\
PNAS + Cutout~\cite{Liu2018_Progressive} & 5 & 3 & 48 & 8 & 17.63 & 3.2M \\
PNAS + Cutout~\cite{Liu2018_Progressive} & 5 & 6 & 128 & 8 & 16.70 & 53.0M \\
ENAS~\cite{Pham2018_ENAS} & 5 & 5 & 36 & 5 & 19.43 & 4.6M \\
ENAS + Cutout~\cite{Liu2018_Progressive} & 5 & 5 & 36 & 5 & 17.27 & 4.6M \\
ENAS + Cutout~\cite{Liu2018_Progressive} & 5 & 5 & 128 & 5 & 16.44 & 52.7M \\
AmoebaNet-B~\cite{Real2019_Aging} & 5 & 6 & 128 & 19 & 17.66 & 34.9M \\
AmoebaNet-B + Cutout~\cite{Real2019_Aging} & 5 & 6 & 128 & 19 & 15.80 & 34.9M \\
NAONet + Cutout~\cite{Luo2018_NAO} & 5 & 6 & 36 & 11 & 15.67 & 10.8M \\
NAONet + Cutout~\cite{Luo2018_NAO} & 5 & 6 & 128 & 11 & 14.75 & 128M \\
\midrule
TAPAS~\cite{Istrate2019_TAPAS} & / & / & / & / & 18.99 & 15.2M \\
Best on source task CIFAR-100 & 5 & 6 & 32 & 19 & 19.96 & 6.3M \\
\midrule
XferNASNet & 5 & 6 & 32 & 19 & 18.88 & 4.5M \\
XferNASNet + Cutout & 5 & 6 & 32 & 19 & 16.29 & 4.5M \\
XferNASNet & 5 & 6 & 64 & 19 & 16.35 & 17.6M \\
XferNASNet + Cutout & 5 & 6 & 64 & 19 & 14.72 & 17.6M \\
XferNASNet & 5 & 6 & 128 & 19 & 15.95 & 69.6M \\
XferNASNet + Cutout & 5 & 6 & 128 & 19 & 14.06 & 69.6M \\
\bottomrule
\end{tabular}
\caption{Various architectures discovered on CIFAR-10 applied to CIFAR-100.
Although the hyperparameters have not been optimized, XferNASNet achieves the best result.
}
\label{tab:experiment-cifar100}
\end{table*}
\subsection{Architecture transfer to CIFAR-100}
A standard procedure to test the transferability is to apply the architectures discovered on CIFAR-10 to another data set, e.g. CIFAR-100~\cite{Luo2018_NAO}.
Although, some of the popular architectures have been adapted to the new data set through additional hyperparameter optimization, we refrain from this for XferNASNet.
We compare the transferability of XferNASNet to other automatically discovered architectures and list the results in Table~\ref{tab:experiment-cifar100}.
For these results, we rely on the numbers reported by Luo et al.~\cite{Luo2018_NAO}.
The XferNASNet architecture achieves an error of 18.88 without and an error of 16.29 with cutout.
Thus, we achieve significantly better results than all other architectures except NAONet.
However, when we increase the number of filters from 32 to 64, we achieve comparable results as NAO with 128 filters and much more parameters.
Furthermore, when we increase the number of filters to 128, the error drops to about 14.06, which is significantly lower than that of NAONet (14.75), and notably with only about half the number of parameters.
We also report the results obtained for the best architecture found during the random search (19.96) in order to reconfirm that XferNAS is discovering better architectures than the ones available for the source tasks.

\begin{figure}[t]
    \centering
    \includegraphics[width=0.5\textwidth]{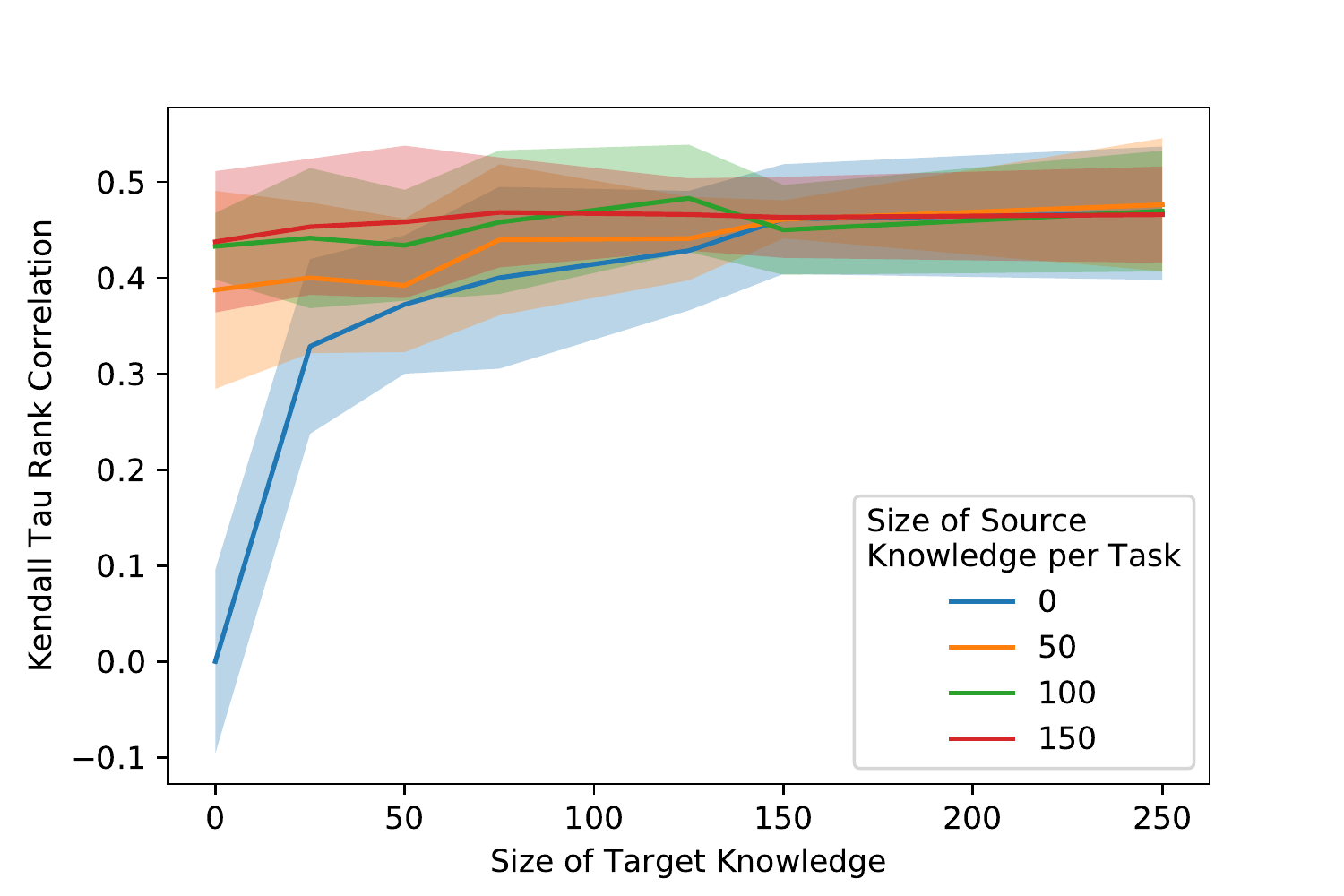}
    \caption{Correlation coefficient between predicted and true validation accuracy with varying amount of source and target knowledge.
     The source knowledge significantly improves predictions when there is little target knowledge available.}
    \label{fig:ablation-study}
\end{figure}
\subsection{Ablation study of XferNet}
At this point we would like to closely examine the benefits of knowledge transfer, especially the circumstances under which we observe a positive effect.
For this we conduct an experiment where we evaluate 600 random architectures for CIFAR-10 according to Section~\ref{sub:source-tasks} and hold out 50 to compute the correlation between predicted and true validation accuracy.
The remaining 550 are candidates for the target knowledge available during the training of the surrogate model.
In this experiment the amount of source and target knowledge is varied.
For every amount of source and target knowledge, ten random splits are created which are used throughout the experiment.
We train the surrogate model as described in Section~\ref{sub:xfernas} on all ten splits and test it on our held-out set.
While the architectures within the observation history are evaluated on exactly one data set, the architectures for evaluation are unknown to the model.
In Figure~\ref{fig:ablation-study}, we visualize the mean and standard deviation of the correlation between the surrogate model prediction and the actual validation accuracy over the ten repetitions.
The x-axis indicates the size of the target knowledge, and the four curves represent experiments corresponding to different sizes of source knowledge.
The source knowledge size ranges from 0 architectures per source task (no knowledge transferred, equivalent to NAO) to 150.
We elaborate on four scenarios in this context.

\textbf{How significant is the benefit of knowledge transfer for a new search (zero target knowledge)?}
This is the scenario in which any method that does not transfer knowledge can not be better than random guessing (correlation of 0).
If our hypothesis is correct and knowledge can be transferred, this should be the scenario in which our method achieves the best results.
And indeed, the correlation is quite high and increases with the amount of source knowledge.

\textbf{Does the transfer model benefit from the target knowledge?}
For any amount of source knowledge, additional target knowledge increases the correlation and, accordingly, improves the predictions.
This effect depends inversely on the amount of source knowledge.

\textbf{What amount of target knowledge is sufficient, so that source knowledge no longer yields a positive effect?}
For target knowledge comprising of 150 architectures (about 30 GPU days), the effect of source knowledge seems to fade away.
Therefore, one can conclude that knowledge transfer does not contribute to any further improvement.

\textbf{When this threshold is reached, does the knowledge transfer harm the model?}
We continue to experiment with larger sizes of target knowledge (up to 550 architectures, not shown in the plot) and empirically confirm that the additional source knowledge does not deteriorate the model performance.
However, the correlation keeps improving with increasing amount of target knowledge for both cases, with and without knowledge transfer.

\section{Conclusions}
In this paper, we present the idea of accelerating NAS by transferring the knowledge about architectures across different tasks.
We develop a transfer framework that extends existing NAS optimizers with this capability and requires minimal modification.
By integrating this framework into NAO, we demonstrate how simple these changes are and also evaluate the resulting new XferNAS optimizer on CIFAR-10 and CIFAR-100.
In just six GPU days, we discover XferNASNet which reaches a new record low for NAS optimizers on CIFAR-10 (2.11$\rightarrow$1.99) and CIFAR-100 (14.75$\rightarrow$14.06) with significantly fewer parameters.

\bibliographystyle{plain}
\bibliography{references}

\cleardoublepage

\appendix
\section{Training details for the convolutional neural networks}
During the search process smaller architectures are trained (B=5, N=3, F=32) for 100 epochs.
The final architecture is trained for 600 epochs according to the specified settings.
SGD with momentum set to 0.9 and cosine schedule~\cite{Loshchilov2017_SGDR} with $l_\text{max}=0.024$ and without warmstart is used for training.
Models are regularized by means of weight decay of $5\cdot 10^{-4}$ and drop-path~\cite{Larsson2017_FractalNet} probability of 0.3.
We use a batch size of 128 but decrease it to 64 for computational reasons for architectures with $F\geq64$.
All experiments use a single V100 graphics card.
The only exception are networks with $F=128$ where we use two V100s to speed up the training process.

Standard preprocessing (whitening) and data augmentation are used.
Images are padded to a size of $40\times 40$ and then randomly cropped to $32\times32$.
Additionally, they are flipped horizontally at random during training.
Whenever cutout~\cite{DeVries2018_Improved} is applied, a size of 16 is used.

\section{Data splits}
The default train/test split of CIFAR-10, CIFAR-100, Fashion-MNIST and SVHN are used and the train split is further divided into 5,000 images for validation and the remaining for training.
For computational reasons we refrain from using the entire Quickdraw data set (50 million drawings and 345 classes).
We select 100 classes at random and select 450 drawings for training and 50 for validation per class at random.

\section{XferNASNet}
\begin{figure*}
    \centering
    \includegraphics[width=1\textwidth]{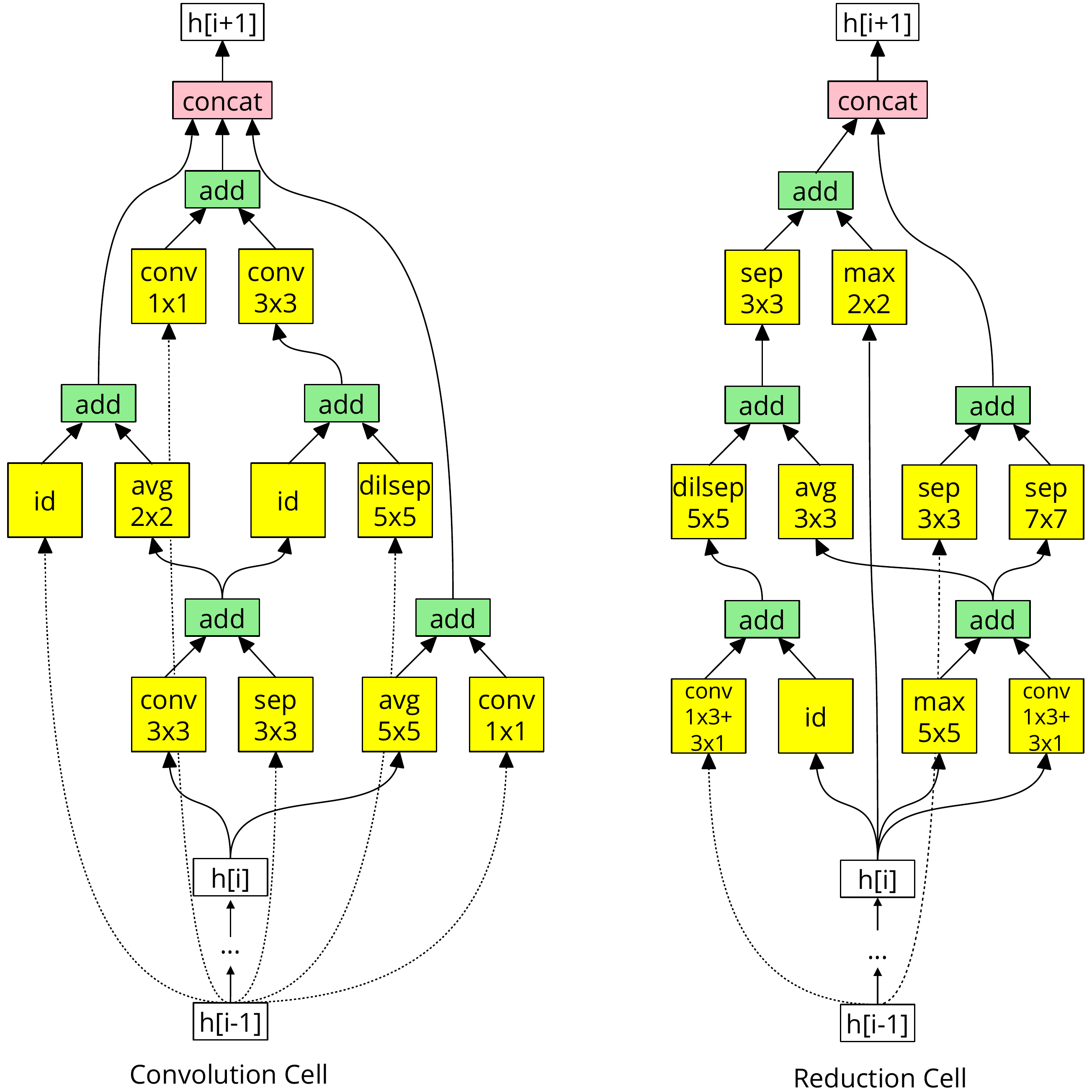}
    \caption{Convolution and reduction cell of XferNASNet.}
    \label{fig:XferNASNet}
\end{figure*}

In this section we provide details about the discovered XferNASNet.
The cells of the architecture are visualized in Figure~\ref{fig:XferNASNet}.
The following json file can be used to reproduce our results using the code found at \url{https://github.com/renqianluo/NAO}.

\begin{verbatim}
{
  "conv_dag": {
    "node_1": ["node_1", null, null, null, null],
    "node_2": ["node_2", null, null, null, null],
    "node_3": ["node_3", "node_1", "node_2", "sep_conv 3x3", "conv 3x3"],
    "node_4": ["node_4", "node_1", "node_3", "identity", "avg_pool 2x2"],
    "node_5": ["node_5", "node_1", "node_3", "dil_sep_conv 5x5", "identity"],
    "node_6": ["node_6", "node_1", "node_2", "conv 1x1", "avg_pool 5x5"],
    "node_7": ["node_7", "node_5", "node_1", "conv 3x3", "conv 1x1"]
  },
  "reduc_dag": {
    "node_1": ["node_1", null, null, null, null],
    "node_2": ["node_2", null, null, null, null],
    "node_3": ["node_3", "node_2", "node_1", "identity", "conv 1x3+3x1"],
    "node_4": ["node_4", "node_2", "node_2", "max_pool 5x5", "conv 1x3+3x1"],
    "node_5": ["node_5", "node_4", "node_3", "avg_pool 3x3", "dil_sep_conv 5x5"],
    "node_6": ["node_6", "node_2", "node_5", "max_pool 2x2", "sep_conv 3x3"],
    "node_7": ["node_7", "node_4", "node_1", "sep_conv 7x7", "sep_conv 3x3"]
  }
}
\end{verbatim}

\end{document}

%% file: xfernet-rl.tikz
\begin{tikzpicture}
\definecolor{colorSegment}{HTML}{ffec9e}
\definecolor{colorCell}{HTML}{ddff88}
\definecolor{colorReductionCell}{HTML}{ffdd88}
\definecolor{colorOp}{HTML}{a6f4fd}
\definecolor{colorBlock}{HTML}{bbee66}

\tikzstyle{boxstyle}=[rectangle,node distance=1.2cm,rounded corners=1ex,align=center]
\tikzstyle{opstyle}=[boxstyle,draw=black,fill=colorOp, minimum size=7mm,minimum width=1.8cm,align=center]
\tikzstyle{concatstyle}=[opstyle]
\tikzstyle{xfernetstyle}=[opstyle,fill=gray!10]
\tikzstyle{segmentstyle}=[opstyle,fill=colorSegment]

\node[segmentstyle](rnn) at (0,0){RNN};
\node[xfernetstyle](xfernet) at (0,1.25){XferNet};
\node[opstyle](softmax) at (0,2.5){softmax};
\node[rectangle](output) at (0,3.5){};

\draw[->]   (rnn) -- (xfernet);
\draw[->]   (xfernet) -- (softmax);
\draw[->]   (softmax) --node[right]{$\pi^{(i)}$} (output);
\end{tikzpicture}

%% file: xfernet-smbo.tikz
\begin{tikzpicture}
\definecolor{colorSegment}{HTML}{ffec9e}
\definecolor{colorCell}{HTML}{ddff88}
\definecolor{colorReductionCell}{HTML}{ffdd88}
\definecolor{colorOp}{HTML}{a6f4fd}
\definecolor{colorBlock}{HTML}{bbee66}

\tikzstyle{boxstyle}=[rectangle,node distance=1.2cm,rounded corners=1ex,align=center]
\tikzstyle{opstyle}=[boxstyle,draw=black,fill=colorOp, minimum size=7mm,minimum width=1.8cm,align=center]
\tikzstyle{concatstyle}=[opstyle]
\tikzstyle{xfernetstyle}=[opstyle,fill=gray!10]
\tikzstyle{segmentstyle}=[opstyle,fill=colorSegment]

\node[rectangle](input) at (0,-1){};
\node[segmentstyle](rnn) at (0,0){RNN};
\node[xfernetstyle](xfernet) at (0,1.25){XferNet};
\node[rectangle](output) at (0,2.25){};

\draw[->]   (input) --node[right]{$a$} (rnn);
\draw[->]   (rnn) -- (xfernet);
\draw[->]   (xfernet) --node[right]{$\hat{f}^{(i)}(a)$} (output);
\end{tikzpicture}

%% file: xfernet.tikz
\begin{tikzpicture}
\definecolor{colorSegment}{HTML}{ffec9e}
\definecolor{colorCell}{HTML}{ddff88}
\definecolor{colorReductionCell}{HTML}{ffdd88}
\definecolor{colorOp}{HTML}{a6f4fd}
\definecolor{colorBlock}{HTML}{bbee66}

\tikzstyle{boxstyle}=[rectangle,node distance=1.2cm,rounded corners=1ex,align=center]
\tikzstyle{opstyle}=[boxstyle,draw=black,fill=colorOp, minimum size=7mm,align=center]
\tikzstyle{concatstyle}=[opstyle]
\tikzstyle{segmentstyle}=[boxstyle,draw=black,fill=colorSegment]
\tikzstyle{cellstyle}=[boxstyle,draw=black,fill=colorCell,align=center]

\xdef\encoderx{1.5}
\xdef\decoderx{9}
\xdef\tasknetsy{2}

\node[opstyle](common) at (3.5,\tasknetsy){Universal\\Network};

\node[rectangle,fill=gray!10,draw=black,dashed,minimum width=5.3cm,minimum height=3cm,align=center,text height=3cm](taskbox) at (7.25,\tasknetsy){\textbf{Residual Task Networks}};
\node[rectangle](taskdots) at (6.95,\tasknetsy){\Large\textbf{...}};

\node[opstyle](task1) at (5.7,\tasknetsy){Residual\\Task Net $1$};

\node[opstyle](taskn) at (8.5,\tasknetsy){Residual\\Task Net $n+1$};

\node[rectangle](archcode) at (5.7,0){};

\node[opstyle](add) at (5.7,\tasknetsy+2.2){add};
\node[rectangle](surrogate) at (5.7,\tasknetsy+3.6){};

\draw[->]   (archcode) -| (common);
\draw[->]   (common) |-node[right,pos=0.2]{$g^{(u)}$} (add);

\draw[->]   (archcode) -- (task1);
\draw[->]   (task1) --node[right]{$r^{(1)}$} (add);

\draw[->]   (archcode) -| (taskn);
\draw[->]   (taskn) |-node[right,pos=0.2]{$r^{(n+1)}$} (add);

\draw[->]   (add) --node[right]{$g^{(i)}$} (surrogate);
\end{tikzpicture}

%% file: optimizer-framework.tikz
\begin{tikzpicture}
\definecolor{colorSegment}{HTML}{ffec9e}
\definecolor{colorCell}{HTML}{ddff88}
\definecolor{colorReductionCell}{HTML}{ffdd88}
\definecolor{colorOp}{HTML}{a6f4fd}
\definecolor{colorBlock}{HTML}{bbee66}

\tikzstyle{boxstyle}=[rectangle,node distance=1.2cm,rounded corners=1ex,align=center]
\tikzstyle{opstyle}=[boxstyle,draw=black,fill=colorOp, minimum size=7mm,align=center]
\tikzstyle{concatstyle}=[opstyle]
\tikzstyle{segmentstyle}=[boxstyle,draw=black,fill=colorSegment]
\tikzstyle{cellstyle}=[boxstyle,draw=black,fill=colorCell,align=center]

\xdef\encoderx{1.5}
\xdef\decoderx{9}

\node[boxstyle,draw=black,fill=gray!10,minimum size=7mm](xfernet) at (6,1.2){XferNet};

\node[rectangle,draw=black,dashed,minimum width=13cm,minimum height=2.5cm,text height=-1.8cm,text width=13cm,fill=gray!0](autoencoderbox) at (6,-0.75){\textbf{Architecture Autoencoder}};
\node[rectangle](input) at (0,-0.75){$a$};
\node[rectangle](output) at (12,-0.75){$\hat{a}$};
\draw[fill=colorSegment] (\encoderx, 0) -- (\encoderx+1,0) -- (\encoderx+1.5,-0.5) -- (\encoderx+1.5,-1) -- (\encoderx+1,-1.5) -- (\encoderx,-1.5) -- cycle;
\node[rectangle] at (\encoderx+0.75,-0.75){Encoder};
\node[cellstyle](archcode) at (6,-0.75){Continuous\\Architecture Code};
\draw[fill=colorSegment] (\decoderx+1.5, 0) -- (\decoderx+0.5,0) -- (\decoderx,-0.5) -- (\decoderx,-1) -- (\decoderx+0.5,-1.5) -- (\decoderx+1.5,-1.5) -- cycle;
\node[rectangle] at (\decoderx+0.75,-0.75){Decoder};

\node[rectangle](surrogate) at (6,2.3){$\hat{f}^{(i)}(a)$};

\draw[->]   (\encoderx+1.5,-0.75) -- (archcode);
\draw[->]   (archcode) -- (\decoderx,-0.75);
\draw[->]   (xfernet) -- (surrogate);
\draw[->]   (input) -- (\encoderx,-0.75);

\draw[->]   (archcode) -- (xfernet);

\draw[->]   (\decoderx+1.5,-0.75) -- (output);
\end{tikzpicture}

%% file: main.bbl
\begin{thebibliography}{10}

\bibitem{Baker2017_Designing}
Bowen Baker, Otkrist Gupta, Nikhil Naik, and Ramesh Raskar.
\newblock Designing neural network architectures using reinforcement learning.
\newblock In {\em 5th International Conference on Learning Representations,
  {ICLR} 2017, Toulon, France, April 24-26, 2017, Conference Track
  Proceedings}, 2017.

\bibitem{Bender2018_Understanding}
Gabriel Bender, Pieter-Jan Kindermans, Barret Zoph, Vijay Vasudevan, and Quoc
  Le.
\newblock Understanding and simplifying one-shot architecture search.
\newblock In Jennifer Dy and Andreas Krause, editors, {\em Proceedings of the
  35th International Conference on Machine Learning}, volume~80 of {\em
  Proceedings of Machine Learning Research}, pages 550--559, Stockholmsmässan,
  Stockholm Sweden, 10--15 Jul 2018. PMLR.

\bibitem{Cai2018_Efficient}
Han Cai, Tianyao Chen, Weinan Zhang, Yong Yu, and Jun Wang.
\newblock Efficient architecture search by network transformation.
\newblock In {\em Proceedings of the Thirty-Second {AAAI} Conference on
  Artificial Intelligence, (AAAI-18), the 30th innovative Applications of
  Artificial Intelligence (IAAI-18), and the 8th {AAAI} Symposium on
  Educational Advances in Artificial Intelligence (EAAI-18), New Orleans,
  Louisiana, USA, February 2-7, 2018}, pages 2787--2794, 2018.

\bibitem{Cai2018_Path}
Han Cai, Jiacheng Yang, Weinan Zhang, Song Han, and Yong Yu.
\newblock Path-level network transformation for efficient architecture search.
\newblock In {\em Proceedings of the 35th International Conference on Machine
  Learning, {ICML} 2018, Stockholmsm{\"{a}}ssan, Stockholm, Sweden, July 10-15,
  2018}, pages 677--686, 2018.

\bibitem{Cai2019_Proxyless}
Han Cai, Ligeng Zhu, and Song Han.
\newblock Proxyless{NAS}: Direct neural architecture search on target task and
  hardware.
\newblock In {\em Proceedings of the International Conference on Learning
  Representations, {ICLR} 2019, New Orleans, Louisiana, USA}, 2019.

\bibitem{DeVries2018_Improved}
Terrance DeVries and Graham~W. Taylor.
\newblock Improved regularization of convolutional neural networks with cutout.
\newblock {\em CoRR}, abs/1708.04552, 2017.

\bibitem{Gastaldi2017_Shake}
Xavier Gastaldi.
\newblock Shake-shake regularization.
\newblock {\em CoRR}, abs/1705.07485, 2017.

\bibitem{Ha2018_Quickdraw}
David Ha and Douglas Eck.
\newblock A neural representation of sketch drawings.
\newblock In {\em 6th International Conference on Learning Representations,
  {ICLR} 2018, Vancouver, BC, Canada, April 30 - May 3, 2018, Conference Track
  Proceedings}, 2018.

\bibitem{Huang2017_Densely}
Gao Huang, Zhuang Liu, Laurens van~der Maaten, and Kilian~Q. Weinberger.
\newblock Densely connected convolutional networks.
\newblock In {\em 2017 {IEEE} Conference on Computer Vision and Pattern
  Recognition, {CVPR} 2017, Honolulu, HI, USA, July 21-26, 2017}, pages
  2261--2269. {IEEE} Computer Society, 2017.

\bibitem{Istrate2019_TAPAS}
Roxana Istrate, Florian Scheidegger, Giovanni Mariani, Dimitrios~S.
  Nikolopoulos, Costas Bekas, and A.~Cristiano~I. Malossi.
\newblock {TAPAS:} train-less accuracy predictor for architecture search.
\newblock In {\em Proceedings of the Thirty-Third {AAAI} Conference on
  Artificial Intelligence, (AAAI-19), Honolulu, Hawaii, USA}, 2019.

\bibitem{Kingma2015_Adam}
Diederik~P. Kingma and Jimmy Ba.
\newblock Adam: {A} method for stochastic optimization.
\newblock In {\em 3rd International Conference on Learning Representations,
  {ICLR} 2015, San Diego, CA, USA, May 7-9, 2015, Conference Track
  Proceedings}, 2015.

\bibitem{Krizhevsky2009_CIFAR}
Alex Krizhevsky.
\newblock Learning multiple layers of features from tiny images.
\newblock Technical report, 2009.

\bibitem{Larsson2017_FractalNet}
Gustav Larsson, Michael Maire, and Gregory Shakhnarovich.
\newblock Fractalnet: Ultra-deep neural networks without residuals.
\newblock In {\em 5th International Conference on Learning Representations,
  {ICLR} 2017, Toulon, France, April 24-26, 2017, Conference Track
  Proceedings}, 2017.

\bibitem{Liu2018_Progressive}
Chenxi Liu, Barret Zoph, Maxim Neumann, Jonathon Shlens, Wei Hua, Li-Jia Li,
  Li~Fei-Fei, Alan Yuille, Jonathan Huang, and Kevin Murphy.
\newblock Progressive neural architecture search.
\newblock In {\em Proceedings of the European Conference on Computer Vision
  (ECCV)}, pages 19--34, 2018.

\bibitem{Liu2018_Hierarchical}
Hanxiao Liu, Karen Simonyan, Oriol Vinyals, Chrisantha Fernando, and Koray
  Kavukcuoglu.
\newblock Hierarchical representations for efficient architecture search.
\newblock In {\em 6th International Conference on Learning Representations,
  {ICLR} 2018, Vancouver, BC, Canada, April 30 - May 3, 2018, Conference Track
  Proceedings}, 2018.

\bibitem{Liu2018_DARTS}
Hanxiao Liu, Karen Simonyan, and Yiming Yang.
\newblock {DARTS:} differentiable architecture search.
\newblock In {\em Proceedings of the International Conference on Learning
  Representations, {ICLR} 2019, New Orleans, Louisiana, USA}, 2019.

\bibitem{Liu2016_SSD}
Wei Liu, Dragomir Anguelov, Dumitru Erhan, Christian Szegedy, Scott~E. Reed,
  Cheng{-}Yang Fu, and Alexander~C. Berg.
\newblock {SSD:} single shot multibox detector.
\newblock In {\em Computer Vision - {ECCV} 2016 - 14th European Conference,
  Amsterdam, The Netherlands, October 11-14, 2016, Proceedings, Part {I}},
  pages 21--37, 2016.

\bibitem{Loshchilov2017_SGDR}
Ilya Loshchilov and Frank Hutter.
\newblock {SGDR:} stochastic gradient descent with warm restarts.
\newblock In {\em 5th International Conference on Learning Representations,
  {ICLR} 2017, Toulon, France, April 24-26, 2017, Conference Track
  Proceedings}, 2017.

\bibitem{Luo2018_NAO}
Renqian Luo, Fei Tian, Tao Qin, Enhong Chen, and Tie{-}Yan Liu.
\newblock Neural architecture optimization.
\newblock In {\em Advances in Neural Information Processing Systems 31: Annual
  Conference on Neural Information Processing Systems 2018, NeurIPS 2018, 3-8
  December 2018, Montr{\'{e}}al, Canada.}, pages 7827--7838, 2018.

\bibitem{Netzer2011_SVHN}
Yuval Netzer, Tao Wang, Adam Coates, Alessandro Bissacco, Bo~Wu, and Andrew~Y.
  Ng.
\newblock Reading digits in natural images with unsupervised feature learning.
\newblock In {\em NIPS Workshop on Deep Learning and Unsupervised Feature
  Learning 2011}, 2011.

\bibitem{Pham2018_ENAS}
Hieu Pham, Melody Guan, Barret Zoph, Quoc Le, and Jeff Dean.
\newblock Efficient neural architecture search via parameters sharing.
\newblock In Jennifer Dy and Andreas Krause, editors, {\em Proceedings of the
  35th International Conference on Machine Learning}, volume~80 of {\em
  Proceedings of Machine Learning Research}, pages 4095--4104,
  Stockholmsmässan, Stockholm Sweden, 10--15 Jul 2018. PMLR.

\bibitem{Real2019_Aging}
Esteban Real, Alok Aggarwal, Yanping Huang, and Quoc~V. Le.
\newblock Aging evolution for image classifier architecture search.
\newblock In {\em Proceedings of the Thirty-Third {AAAI} Conference on
  Artificial Intelligence, (AAAI-19), Honolulu, Hawaii, USA}, 2019.

\bibitem{Real2017_Large}
Esteban Real, Sherry Moore, Andrew Selle, Saurabh Saxena, Yutaka~Leon Suematsu,
  Jie Tan, Quoc~V. Le, and Alexey Kurakin.
\newblock Large-scale evolution of image classifiers.
\newblock In Doina Precup and Yee~Whye Teh, editors, {\em Proceedings of the
  34th International Conference on Machine Learning}, volume~70 of {\em
  Proceedings of Machine Learning Research}, pages 2902--2911, International
  Convention Centre, Sydney, Australia, 06--11 Aug 2017. PMLR.

\bibitem{Redmon2016_YOLO}
Joseph Redmon, Santosh~Kumar Divvala, Ross~B. Girshick, and Ali Farhadi.
\newblock You only look once: Unified, real-time object detection.
\newblock In {\em 2016 {IEEE} Conference on Computer Vision and Pattern
  Recognition, {CVPR} 2016, Las Vegas, NV, USA, June 27-30, 2016}, pages
  779--788, 2016.

\bibitem{Wistuba2018_Deep}
Martin Wistuba.
\newblock Deep learning architecture search by neuro-cell-based evolution with
  function-preserving mutations.
\newblock In {\em {ECML/PKDD} {(2)}}, volume 11052 of {\em Lecture Notes in
  Computer Science}, pages 243--258. Springer, 2018.

\bibitem{Wistuba2018_Practical}
Martin Wistuba.
\newblock Practical deep learning architecture optimization.
\newblock In {\em 5th {IEEE} International Conference on Data Science and
  Advanced Analytics, {DSAA} 2018, Turin, Italy, October 1-3, 2018}, pages
  263--272, 2018.

\bibitem{Wistuba2019_Inductive}
Martin Wistuba and Tejaswini Pedapati.
\newblock Inductive transfer for neural architecture optimization.
\newblock {\em CoRR}, abs/1903.03536, 2019.

\bibitem{Wistuba2019_A}
Martin Wistuba, Ambrish Rawat, and Tejaswini Pedapati.
\newblock A survey on neural architecture search.
\newblock {\em CoRR}, abs/1905.01392, 2019.

\bibitem{Wong2018_Transfer}
Catherine Wong, Neil Houlsby, Yifeng Lu, and Andrea Gesmundo.
\newblock Transfer learning with neural automl.
\newblock In {\em Advances in Neural Information Processing Systems 31: Annual
  Conference on Neural Information Processing Systems 2018, NeurIPS 2018, 3-8
  December 2018, Montr{\'{e}}al, Canada.}, pages 8366--8375, 2018.

\bibitem{Xiao2017_Fashion-MNIST}
Han Xiao, Kashif Rasul, and Roland Vollgraf.
\newblock Fashion-mnist: a novel image dataset for benchmarking machine
  learning algorithms.
\newblock {\em CoRR}, abs/1708.07747, 2017.

\bibitem{Xie2017_Aggregated}
Saining Xie, Ross~B. Girshick, Piotr Doll{\'{a}}r, Zhuowen Tu, and Kaiming He.
\newblock Aggregated residual transformations for deep neural networks.
\newblock In {\em 2017 {IEEE} Conference on Computer Vision and Pattern
  Recognition, {CVPR} 2017, Honolulu, HI, USA, July 21-26, 2017}, pages
  5987--5995, 2017.

\bibitem{Xie2019_SNAS}
Sirui Xie, Hehui Zheng, Chunxiao Liu, and Liang Lin.
\newblock {SNAS}: stochastic neural architecture search.
\newblock In {\em Proceedings of the International Conference on Learning
  Representations, {ICLR} 2019, New Orleans, Louisiana, USA}, 2019.

\bibitem{Zhong2018_Practical}
Zhao Zhong, Junjie Yan, Wei Wu, Jing Shao, and Cheng{-}Lin Liu.
\newblock Practical block-wise neural network architecture generation.
\newblock In {\em 2018 {IEEE} Conference on Computer Vision and Pattern
  Recognition, {CVPR} 2018, Salt Lake City, UT, USA, June 18-22, 2018}, pages
  2423--2432, 2018.

\bibitem{Zoph2017_Neural}
Barret Zoph and Quoc~V. Le.
\newblock Neural architecture search with reinforcement learning.
\newblock In {\em 5th International Conference on Learning Representations,
  {ICLR} 2017, Toulon, France, April 24-26, 2017, Conference Track
  Proceedings}, 2017.

\bibitem{Zoph2018_Learning}
Barret Zoph, Vijay Vasudevan, Jonathon Shlens, and Quoc~V. Le.
\newblock Learning transferable architectures for scalable image recognition.
\newblock In {\em 2018 {IEEE} Conference on Computer Vision and Pattern
  Recognition, {CVPR} 2018, Salt Lake City, UT, USA, June 18-22, 2018}, pages
  8697--8710, 2018.

\end{thebibliography}
